\documentclass[sigconf]{acmart}
\usepackage{epsfig}
\usepackage{graphicx}
\usepackage{amsmath}

\usepackage{amssymb}

\def\ie{\textit{i.e.}}

\def\BibTeX{{\rm B\kern-.05em{\sc i\kern-.025em
			b}\kern-.08emT\kern-.1667em\lower.7ex\hbox{E}\kern-.125emX}}

\AtBeginDocument{%
	\providecommand\BibTeX{{%
			\normalfont B\kern-0.5em{\scshape i\kern-0.25em b}\kern-0.8em\TeX}}}

\acmSubmissionID{2017}

\copyrightyear{2020}
\acmYear{2020}
\setcopyright{acmcopyright}\acmConference[MM '20]{Proceedings of the 28th ACM
	International Conference on Multimedia}{October 12--16, 2020}{Seattle, WA, USA}
\acmBooktitle{Proceedings of the 28th ACM International Conference on Multimedia
	(MM '20), October 12--16, 2020, Seattle, WA, USA}
\acmPrice{15.00}
\acmDOI{10.1145/3394171.3413502}
\acmISBN{978-1-4503-7988-5/20/10}

\begin{document}
	
	\fancyhead{} 
	
	\title{Depth Guided Adaptive Meta-Fusion Network \\ for Few-shot Video Recognition}
	
	\author[Y. Fu, L. Zhang, J. Wang, Y. Fu, Y.-G. Jiang]{Yuqian Fu$^{1*}$, Li Zhang$^{2*}$, Junke Wang$^{1}$, Yanwei Fu$^{3}$, Yu-Gang Jiang$^{1\#}$}
	\affiliation{$^1$Shanghai Key Lab of Intelligent Information Processing, School of Computer Science, Fudan University}
	\affiliation{$^2$Department of Engineering Science, University of Oxford}
	\affiliation{$^3$School of Data Science, Fudan University}
	\affiliation{\{yqfu18, 17300240009, yanweifu, ygj\}@fudan.edu.cn, lz@robots.ox.ac.uk}
	\thanks{* indicates equal contributions, $\#$ indicates corresponding author}
	
	\renewcommand{\shortauthors}{Fu and Zhang, et al.}

	\begin{abstract}
		Humans can easily recognize actions with only a few examples given, while the existing video recognition models still heavily rely on the large-scale labeled data inputs. This observation has motivated an increasing interest in few-shot video action recognition, which aims at learning new actions with only very few labeled samples. In this paper, we propose a \emph{depth guided Adaptive Meta-Fusion Network}  for few-shot video recognition which is termed as \emph{AMeFu-Net}. Concretely, we tackle the few-shot recognition problem from three aspects: firstly, we alleviate this extremely data-scarce problem by introducing depth information as a carrier of the scene, which will bring extra visual information to our model; secondly, we fuse the representation of original RGB clips with multiple non-strictly corresponding depth clips sampled by our \emph{temporal asynchronization augmentation} mechanism, which synthesizes new instances at feature-level; {thirdly,} a novel \emph{Depth Guided Adaptive Instance Normalization (DGAdaIN)} fusion module is proposed to fuse the two-stream modalities efficiently. Additionally, to better mimic the few-shot recognition process, our model is trained in the \emph{meta-learning} way. Extensive experiments on several action recognition benchmarks demonstrate the effectiveness of our model.
	\end{abstract}
	
	
	\settopmatter{printacmref=false, printfolios=false}

	\begin{teaserfigure}
	    \centering
		\includegraphics[width=0.9\textwidth]{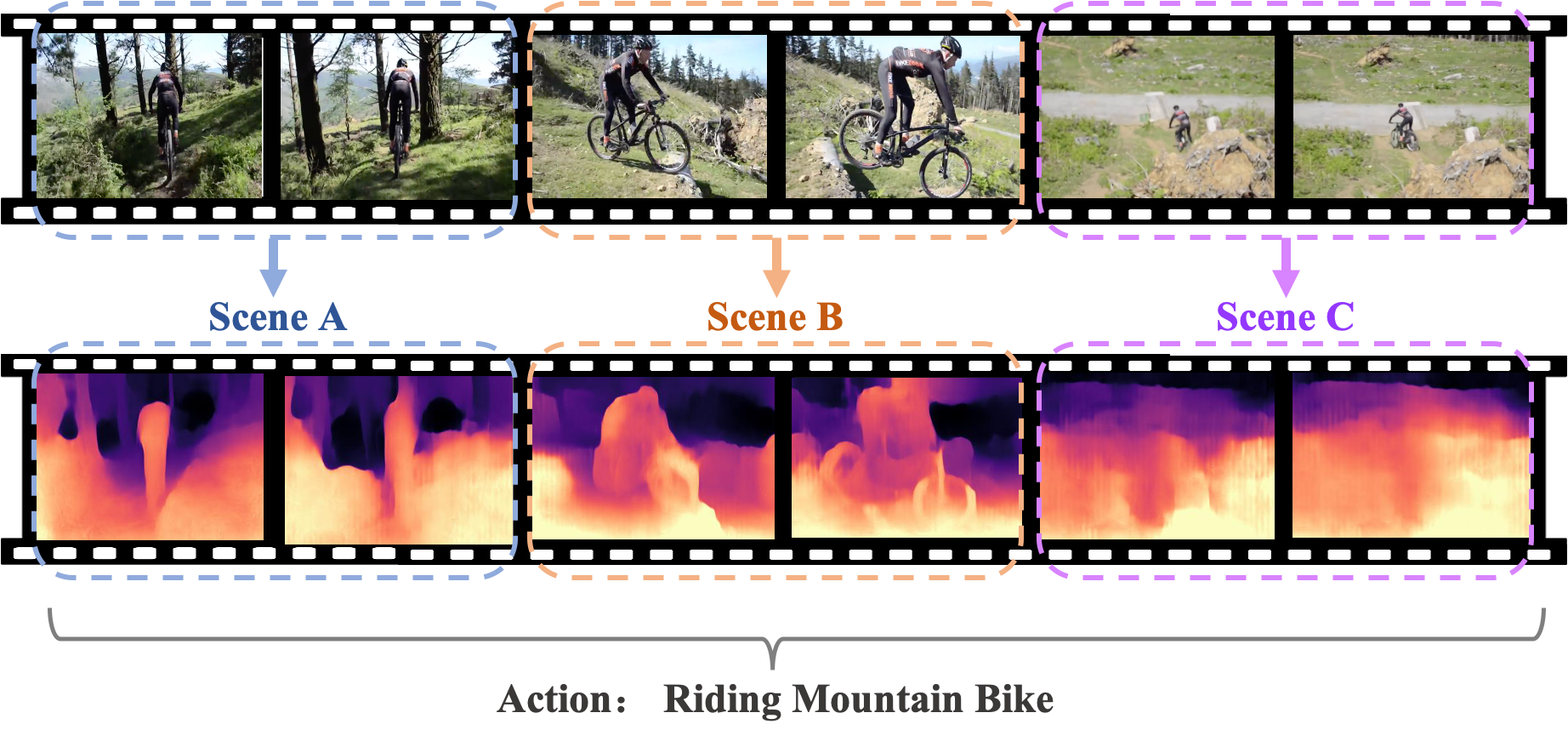}
		\caption{
			\textbf{Concept}: depth information helps to model the relation between the moving person and the scene, thus guiding to learn richer context and less biased representation. We observe two important characteristics: 1) Apparently, scene information helps us to recognize actions. In this example, the mountain is critical to identifying the action as "riding mountain bike". 2) Even if the scene shifts from the mountain to the roadside, we can still recognize it correctly. The example is sampled from the Kinetics dataset. \label{idea} 
		}
		\label{fig:teaser}
	\end{teaserfigure}

	\maketitle
	\vspace{-0.1in}
    {\fontsize{8pt}{8pt} \selectfont\textbf{ACM Reference Format:}\\Yuqian Fu, Li Zhang, Junke Wang, Yanwei Fu, Yu-Gang Jiang. 2020. Depth Guided Adaptive Meta-Fusion Network for Few-shot Video Recognition. In \textit{Proceedings of the 28th ACM International Conference on Multimedia (MM’20), October 12--16, 2020, Seattle, WA, USA.} ACM, New York, NY, USA, 10 pages. https://doi.org/10.1145/3394171.3413502 }
	\vspace{0.1in}
	
	\section{Introduction}
	The ubiquitous availability and use of mobile devices facilitate capturing and sharing videos on social platforms. Therefore, general video understanding is extremely important to real-world multimedia applications, e.g. robotic interaction \cite{xu2019shared} and auto-driving~\cite{geiger2012we}, and thus attracting extensive research attention. Particularly, with the advent of state-of-the-art deep architectures ~\cite{simonyan2014two,carreira2017quo,feichtenhofer2019slowfast,perez2020egocentric}, the performance of video recognition has been dramatically improved in the standard supervised-learning setting. However, a large amount of manually labeled video data are necessary, which is an ideal and yet impractical requirement. In contrast, humans can easily learn a novel concept, even when we have seen only one or extremely few samples. Such a gap inspires the study of few-shot learning in the Multimedia community.

	Generally, it still presents significant challenges for the community to enable models to own the ability to recognize novel classes by learning from limited labeled data, \ie, few-shot learning \cite{siamese_1shot, matchingnet_1shot,relation_net}, as deep models are prone to overfitting with few training instances. This will worsen the generalization ability of the models.
	Most few-shot recognition works have been explored for images~\cite{chen2018image,compositional_1shot,prototype_network,relation_net,wang2020instance,wang2020trust}, but to a lesser extent for videos
	\cite{zhu2018compound,zhang2020few}. Particularly,  learning representations from the temporal sequences of video frames is intrinsically much more difficult and complex than those from images. It is non-trivial to learn good representations for few-shot video recognition. Recent few-shot video recognition efforts have been made on using key-value memory network paradigm~\cite{zhu2018compound}, matching videos via temporal alignment~\cite{cao2019few}, learning from web data~\cite{tu2019learning}, and applying attention mechanism~\cite{zhang2020few}.

	Different from the aforementioned methods, this paper tackles few-shot action recognition from a fundamentally different perspective. 
	We argue that depth modality is useful for facilitating few-shot video action recognition since it is able to model the geometric relations within the context. Besides, shifting depth of a video can implicitly represent the motion information of the objects in the scene. Based on these two insights, in this paper, we explore the depth information guided few-shot video action recognition.
	Specifically, the representation bias~\cite{resound}, \textit{e.g.}, scene representation bias, is inevitably learned by supervised video recognition methods, since human actions often happen in specific scene contexts. Despite the fact that such bias may damage the generalization ability of learned representation~\cite{dance_in_the_mall},  the co-occurrence of actions and corresponding scenes may help to alleviate the difficulty of lack of labeled instances.  To this end, rather than using segmentation methods to directly recognize the scenes, we resort to predicting depth frames to help understand the geometric relations within the scenes as the richer representation of moving persons and their corresponding environments. For example, as shown in Fig.~\ref{idea}, the action of "Riding Mountain Bike" mostly happens in the scene of the mountain, where the geometric relation from corresponding depth images between the person and scene, would essentially help to recognize the action. 
	
	Furthermore, as shown in Fig.~\ref{idea}, we can still recognize the action correctly even if the scene shifts from the mountain (Scene B) to the roadside (Scene C).  
	That is, even though the depth clip is not exactly corresponding to the RGB clip, the model should still be able to recognize the action correctly in principle. 
	Such an asynchronization between depth and RGB clips inspires us to propose a more natural way of augmenting video representations -- temporal asynchronization augmentation mechanism by randomly sampling unpaired RGB and depth clips.
	
	Formally, this paper proposes a novel depth guided Adaptive Meta-Fusion Network (AMeFu-Net) for few-shot video recognition. Specifically, the depth frames are predicted by an off-the-shelf depth predictor \cite{godard2019digging} and serve as a carrier in understanding the geometric relations within the scene. Furthermore, we augment our original videos by sampling some non-strictly corresponding RGB and depth pairs as the training data, which helps us to enhance the robustness of our model. Besides, our model includes a key component, depth guided adaptive instance normalization (DGAdaIN) module, which effectively fuses RGB and depth features. We learn the affine parameters from depth feature adaptively, and then apply them to deform the original RGB feature. Therefore, the information of RGB modality and depth modality are integrated more effectively.
	
	Our model is trained via the meta-learning mechanism~\cite{matchingnet_1shot}. Specifically, it automatically learns cross-task knowledge so that the model can quickly acquire task-specific knowledge of new few-shot learning tasks at inference time. 

	Our contributions are summarized as follows. 
	1) For the first time, we propose to use depth information to alleviate the data-scarce problem in few-shot video action recognition.
	2) We propose a novel temporal asynchronization augmentation mechanism, which randomly samples depth data for RGB data to augment the source video representation.
	3) We propose a depth guided adaptive instance normalization module (DGAdaIN) to better fuse the two-stream features. 
	4) Extensive experiments show that our proposed AMeFu-Net allows us to establish new state-of-the-art on three widely used datasets including Kinetics, UCF101, and HMDB51.

	\section{Related work}
	
	\noindent \textbf{Few-shot learning.}
	Few-shot learning aims at recognizing unseen concepts with only a few labeled samples. 
	Many works have been done to address this problem in the image domain. 
	Flagship works include metric-learning based methods \cite{snell2017prototypical,matchingnet_1shot,liu2020embarrassingly,sung2017learning,zhang2017learning},
	meta-learning methods \cite{finn2017model, wang2017learning, wang2016learningto}, 
	and generative models \cite{Lake_oneshot, miller2000oneshot_transforms,wang2018low}. 
	Comparatively, the video domain still remains under-explored.  \cite{zhu2018compound} proposes a compound key-value memory network paradigm. 
	\cite{fu2019embodied} learns from virtual actions which are generated by embodied agents. 
	\cite{cao2019few} uses a temporal alignment method to compare the similarity between videos. 
	\cite{zhang2020few} handles this problem by introducing an attention mechanism with self-supervised training.
	In this paper, we mainly explore how to alleviate the problem of extremely limited samples in few-shot learning by fusing extra modality information.

	\noindent \textbf{Context for video understanding.} 
	Visual context information has been validated to be very useful for video understanding tasks. Many visual concepts have been explored as a supplementary information of videos. 
	Previous works mainly focus on scenes \cite{li2007and, marszalek2009actions, cvpr16OSS}, 
	objects \cite{gupta2007objects, ikizler2010object, jain201515, wu2007scalable, wang2018videos}, 
	poses or skeletons \cite{jiang2015informative, wang2013approach, yang2010recognizing, yan2019pa3d,yan2018spatial}. 
	However, these methods are prohibitively expensive and sensitive to the noise.
	On the other hand, the middle-level representations, such as objects and the human body skeletons, are too abstract to contain enough information of motion and scenes. 
	Comparatively, the information of depth makes the best of both worlds: it can not only effectively represent the detailed information of objects, scenes, and their potential motion expressed in videos, but also depth information is well robust to noise thanks to recently developed models. 
	To the best of our knowledge, we are the first to introduce depth as the carrier of scene information for video recognition under few-shot learning. 
	Note that \cite{boukhers2018example} estimates the depth to correct the camera odometry.
	Most of the multi-modality models simply fuse the features from different streams by averaging \cite{simonyan2014two}, concatenation \cite{wang2018videos}, recurrent neural networks~\cite{zhang2017actor} or with fully connected layers \cite{ijcai18:face}, while we adaptively learn how to combine the RGB stream and depth stream by introducing a novel depth guided adaptive instance normalization module.

	\noindent \textbf{Instance normalization.} 
	The instance normalization \cite{ulyanov2017improved} is first proposed for feed-forward style-transfer \cite{gatys2016image, huang2017arbitrary} by replacing the widely used Batch Normalization \cite{ioffe2015batch}, resulting in significant improvement to the quality of image stylization. 
	After that, several related methods have been proposed \cite{huang2017arbitrary,Huang_2018_ECCV,Park_2019_CVPR,qian2020long}. 
	Among them, AdaIN \cite{huang2017arbitrary}, a module for style transfer, proposes to align the mean and variance of the content feature of image $A$, with those of the style feature of image $B$. 
	This inspires us to fuse the RGB stream and depth stream by learning the scale and shift affine parameters from depth, thus deforming RGB information towards depth for few-shot video action recognition.

	\noindent \textbf{Temporal shift mechanism.}
	Temporal Shift Module (TSM) \cite{lin2019tsm} utilizes a temporal shift mechanism to shift part of the channels along the temporal dimension to model the spatial-temporal representation of videos. LGTSM \cite{chang2019learnable} further extends the TSM by shifting temporal kernels for video inpainting.  
	Different from previous works which aim at fusing temporal information by shifting features, in this paper, the temporal shift is used to sample asynchronized depth and RGB clips in the temporal dimension to synthesize new instances in feature space.
	By doing this, our method essentially samples unpaired data to generate more abundant video representations. It is fundamentally different to previous data augmentation works~\cite{chen2019image, chen2019imageMeta, fu2019embodied,zhong2017random,hariharan2017low, wang2018low, schwartz2018delta, gao2018low}.

	
	\section{Methodology}
	
	\noindent  \textbf{Problem setup.} 
	For few-shot learning setting, there is a base action set $\mathcal{D}_{base}$ and a novel action set $\mathcal{D}_{novel}$. 
	The categories of the base and novel set are disjoint, that is,  $\mathcal{C}_{base} \cap  \mathcal{C}_{novel} = \emptyset $. 
	All the videos contained in base set  $\mathcal{D}_{base} = \{(V_{i}, c_{i}), c_{i} \in \mathcal{C}_{base}\}$, are used as source data to train the model.
	We denote the $i-$th video and its corresponding class label as $V_{i}$, $c_{i}$, respectively. 
	In few-shot video recognition setting, the goal of our algorithms is to recognize the novel videos contained in novel dataset $\mathcal{D}_{novel} = \{(V_{i}, c_{i}), c_{i} \in \mathcal{C}_{novel}\}$, the recognition model learned on  $\mathcal{D}_{base} $ should generalize to the novel categories contained in $\mathcal{D}_{novel}$, that is, it should be able to recognize $\mathcal{C}_{novel}$ given only one or few labeled example videos per class.

\begin{figure*}[h!]
		\centering
		{\includegraphics[width=0.9\linewidth]{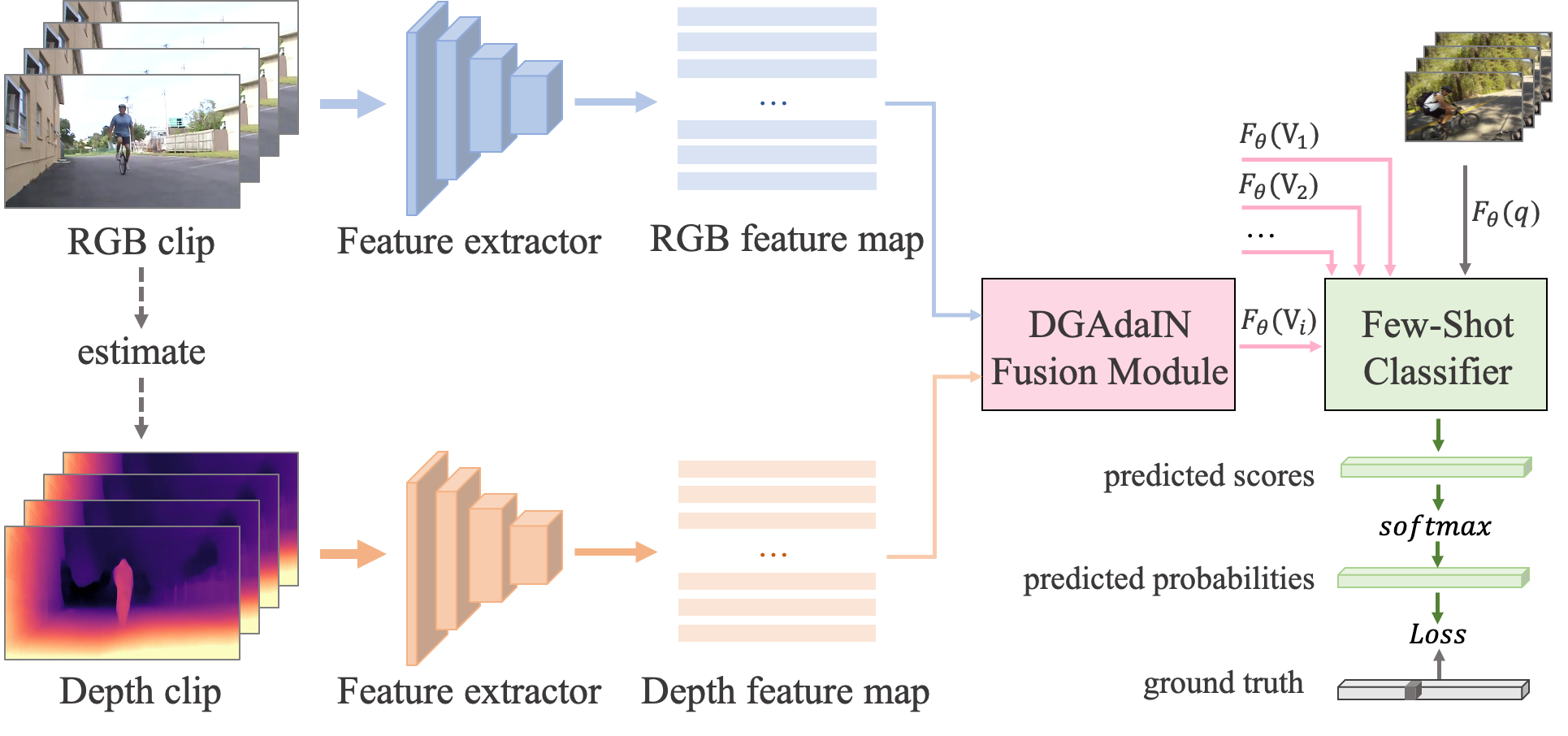}}
		\caption{The schematic illustration of our few-shot video action recognition model. For each video contained in the episode, the $\left \langle RGB\ clip,\ depth\ clip \right \rangle$ pair is first sampled by our \emph{temporal asynchronization augmentation mechanism} as the input to our model. Then two feature extractors are used to extract the feature maps. After that, an \emph{depth guided adaptive instance normalization (DGAdaIN) fusion module} is applied to fuse the feature maps. Finally, all the fused features of videos in support set and query set are fed into our few-shot classifier to predict the action of the query video.}
		\label{framework} 
	\end{figure*}

	\noindent \textbf{Framework overview.} 
	Our proposed AMeFu-Net mainly consists of the following components: 
	backbone with depth, temporal asynchronization augmentation module, depth guided adaptive instance normalization fusion (DGAdaIN) module, and the few-shot classifier. The schematic illustration is shown in Fig.~\ref{framework}.

	Following \cite{zhu2018compound}, we use the meta-learning strategy for model training and testing. Meta-learning~\cite{finn2017model} consists of the meta-train and meta-test phases. Generally, meta-learning can efficiently learn cross-task knowledge with $\mathcal{D}_{base}$ in meta-train phase, so that the model can quickly acquire task-specific knowledge of novel few-shot learning tasks in $\mathcal{D}_{novel}$ at meta-test time. Specifically, we adopt the \emph{episode-based} training strategy  \cite{zhu2018compound,snell2017prototypical,sung2017learning} which randomly sample training instances from $\mathcal{D}_{base}$ for each batch in meta-train phase. Each episode consists of a support set and a query set, and the task of our model is to predict the action class of the query video based on the videos in the corresponding support set. In a \emph{N-way-K-shot} problem, for each episode, a collection of $N$ classes each with $K$ videos are randomly sampled as the support set, and other videos belong to $N$ classes are formed as the query set. 
	Following CMN~\cite{zhu2018compound}, only one video is sampled for the query set in each episode.
	
	During the meta-train phase, we first use the temporal asynchronization augmentation module to sample $\left \langle RGB\ clip,\ depth\ clip \right \rangle$ pairs. The sampled pairs are then fed into the feature extractors to get RGB and depth features, respectively. Then the DGAdaIN adaptive fusion module is adopted to integrate the two-stream information to get the final representations of videos. By virtue of such a way, for the $i$-th training video $V_{i}$ in the support set and the only query video $q$ in the query set, we obtain the corresponding fused feature denoted as $ F_{\theta}(V_{i}) $, and $F_{\theta}(q) $, respectively. Finally, a few-shot classifier is used to predict the probability by comparing the distance between $ F_{\theta}(V_{i})$ and $F_{\theta}(q)$. Concretely, We apply a softmax layer on distance similarities to obtain the predicted probabilities. Finally, cross entropy is used to calculate the loss, thus optimizing our network.
	
	During the meta-test phase, we sample episodes from $\mathcal{D}_{novel}$.
	Different from the meta-train phase, we do not apply the temporal asynchronization augmentation mechanism, which means only one strictly corresponding $\left \langle RGB\ clip,\ depth\ clip \right \rangle$ pair for each video is sampled. The action class with the highest probability is selected as the predicted class.

	\subsection{Backbone with depth}\label{backbone}
	As shown in Fig.~\ref{framework}, the backbone is designed as the two-stream architecture, of which the RGB stream extracts more generic and robust visual information of the input videos, while the depth stream is introduced as supplementary scene information to make up for the lack of data and represent context features. In other words, we address the few-shot problem from a multi-modal perspective. For both streams, we adopt a ResNet-50 \cite{he2016deep} network pre-trained on ImageNet \cite{deng2009imagenet} as the backbone. The output of the last convolution layer is extracted as the representation of videos. We finetune the feature extractor on the source domain to better fit them on the target datasets.

	Our depth maps are obtained by utilizing the off-the-shelf depth estimation model -- Monodepth2 \cite{godard2019digging}. 
	More specifically, it takes a single RGB image as input and predicts its corresponding depth frame. In this work, we take the depth model as an off-the-shelf depth feature extractor, which is pre-trained on KITTI dataset~\cite{geiger2012we}.  Notably, despite the large domain discrepancy between the video datasets used in our experiments and KITTI, the pre-trained depth model can effectively predict a usable depth image,  thus working as a generic depth predictor in our model.

	\subsection{Depth guided adaptive fusion module}
	
	\begin{figure*}[!h]
		\centering
		{\includegraphics[width=0.98\linewidth]{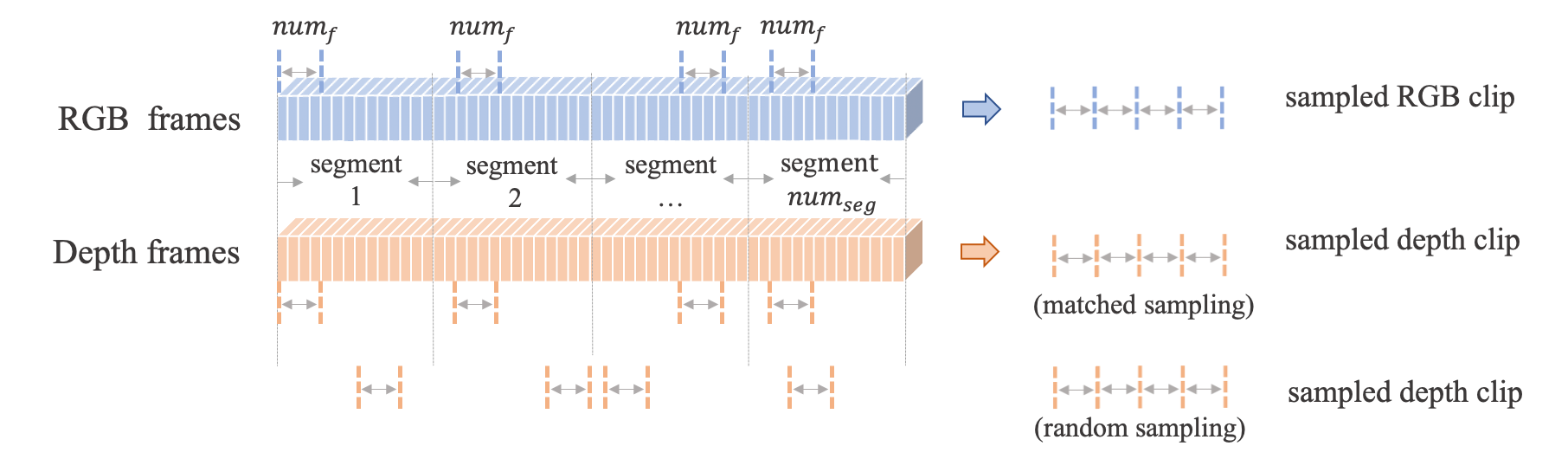}}
		\caption{Illustration of our temporal asynchronization augmentation mechanism. For each modality, we divide the video frames into $num_{seg}$ segments of equal length, then we randomly sample $num_{f}$ consecutive frames from each segment to form the video clip.  More importantly, for the sampled RGB clip, both the matched depth clip and non-strictly matched depth clips are sampled to augment the source dataset.}
		\label{shift} 
	\end{figure*}
	
	In this section, the RGB information and depth information should be combined to facilitate few-shot video recognition. For this purpose, we propose a depth guided adaptive instance normalization (DGAdaIN) fusion module, inspired by the Instance Normalization (IN), which is defined as,
	\begin{equation}
	IN(x) = \gamma \cdot \frac{x-\mathrm{\mu}(x)}{\mathrm{\sigma}(x)} + \beta 
	\end{equation}
	\noindent where $x$ denotes the input batch $x\in \mathbb{R}^{B \times C \times H \times W} $,  $ \mathrm{\mu}(x), \mathrm{\sigma}(x) \in \mathbb{R}^{B \times C} $ are the mean and standard deviation of input $x$, while $\gamma, \beta \in \mathbb{R}^{B \times C} $ are affine parameters learned from data. 
	$B$, $C$, $H$, $W$ denotes the batch size,  channel of feature map (\ie the color channel of an RGB image),  the height and width of the image feature map, respectively.

	Different from the widely used Batch Normalization, where the mean and standard deviation are calculated for each individual feature channel, IN calculates them for each sample; so the $\mathrm{\mu}(x) , \mathrm{\gamma}(x)  $ are computed across the spatial dimensions. Formally,  let $x_{b,c,h,w} $ denotes the {$b,c,h,w-th$} element of the input $x$,
	\begin{equation}
	\mathrm{\mu_{b,c}}(x) = \frac{1}{HW} \sum_{h=1}^{H} \sum_{w=1}^{W} x_{b,c,h,w}
	\end{equation}
	\begin{equation}
	\mathrm{\sigma_{b,c}}(x) = \sqrt{ \frac{1}{HW} \sum_{h=1}^{H} \sum_{w=1}^{W} (x_{b,c,h,w} - \mathrm{\mu_{b,c}}(x))^2 + \epsilon }
	\end{equation}
	
	In our model, the input batch is denoted as $ x \in \mathbb{R}^{B \times { D} \times L}$, where $D$ denotes the number of frames of a single video and $L$ denotes the dimension of feature for each frame.
	
	Inspired by the remarkable success of IN in style transfer, we deform the RGB feature towards the depth to fuse multi-modality features. 
	Therefore, we propose a novel depth guided instance normalization method, in which the $\gamma$ and $\beta$ are designed to learn from the depth feature map adaptively.
	More specifically, our DGAdaIN module $ f\left(\mathrm{I}_{rgb},~\mathrm{I}_{d}\right)$ receives a RGB input batch  $ \mathrm{I}_{rgb} \in \mathbb{R}^{B \times D \times L}$ and a depth input batch  $ \mathrm{I}_{d} \in \mathbb{R}^{B \times D \times L}$,  and learns the affine parameters from the depth stream,
	\begin{equation}\label{eq1} 
	f\left(\mathrm{I}_{rgb},\mathrm{I}_{d}\right)=g_{s}\left(\mathrm{I}_{d}\right)\cdot\frac{\mathrm{I}_{rgb}-\mathrm{\mu}(\mathrm{I}_{rgb})}{\mathrm{\sigma}(\mathrm{I}_{rgb})}+g_{b}\left(\mathrm{I}_{d}\right)
	\end{equation}
	\noindent where $g_s$ and $g_b$  
	are both a learnable fully-connected (FC) layer and jointly optimized with the whole network. 
	The outputs are treated as the affine parameters $\gamma$ and $\beta$ respectively. 
	They are utilized as "scale" and "shift" factors to deform the RGB feature. 
	$\mu_{b,d}$ and $\sigma_{b,d}$ are computed along the $L$ dimension:
	\begin{equation}
	\mathrm{\mu_{b,d}}(I_{rgb}) = \frac{1}{L} \sum_{l=1}^{L}{I_{rgb}}_{b,d,l}
	\end{equation}
	\begin{equation}
	\mathrm{\sigma_{b,d}}(I_{rgb}) = \sqrt{ \frac{1}{L} \sum_{l=1}^{L} ({I_{rgb}}_{b,d,l} - \mathrm{\mu_{b,d}}(I_{rgb}))^2 + \epsilon }
	\end{equation}

	During the meta-train phase, we obtain the fused feature (Eq.~\ref{eq1}) for the query video and support videos. 
	The classification probabilities are then calculated using our few-shot classifier described in section~\ref{sec:classifier}.
	After that, we use the cross entropy loss to update the network.

	\subsection{Temporal asynchronization augmentation \label{section:shift}}
	This module introduces the temporal asynchronization augmentation mechanism for data augmentation. 
	Although the context information facilitates to understand a video, a robust model should have the ability to recognize the video action correctly in different contexts. 
	That is even when the depth clip is unmatched with the RGB clip, our model should still recognize the action. Therefore, we propose to shift the position of depth clip with respect to its corresponding RGB clip along the temporal dimension, as illustrated in Fig.~\ref{shift}.

	Inspired by TSN~\cite{wang2016temporal}, we propose a \textbf{basic sampling strategy} to handle the video inputs for our few-shot action recognition.	Specifically, we first divide the video frames into $num_{seg}$ segments of equal length, then $num_{f}$ consecutive frames are sampled from each segment to form a sampled clip. As for the \textbf{temporal asynchronization augmentation mechanism}, we not only sample the corresponding $\left \langle RGB\ clip,\ depth\ clip \right \rangle$ pair as most multi-modality methods do, but also sample another $num_{aug}$ pairs by randomly select the $num_{f}$ consecutive frames from each segment. During the meta-train phase, the $1 + num_{aug}$ pairs are iteratively used to train our model.  By virtue of this way, we augment our original datasets by $num_{aug}$ times, which enhances the robustness of our model effectively.

	\subsection{Few-shot classifier}\label{sec:classifier} 
	Our few-shot classifier is derived from ProtoNet \cite{snell2017prototypical}.
	Specifically, $S_{c}$ denotes the set of videos with label $c$. 
	For each action $c$, its prototype $p_c$ is calculated by averaging the feature of videos belongs to $S_{c}$ as follows,
	
	\begin{equation}
	p_c = \frac{1}{|S_c|} \sum_{(V_i, c_i) \in {S_c}}F_{\theta}(V_i).
	\end{equation}
	
	Then we predict the action of query video $q$ by comparing the feature of $q$ with the prototypes of different actions. We apply a softmax over the distance similarities to obtain the probabilities of predicted results. Formally, the 
	probability that query video $q$ belongs to action $c$, is defined as follows:
	\begin{equation}
	P_{\theta}(c_{pre} = c | q ) = \frac{exp (||F_\theta(q), p_c||)}{\sum_{i=1}^N exp (||F_\theta(q), p_i)||}
	\end{equation}
	\noindent where $||\ \cdot ||$ denotes the distance similarity, $c_{pre} $ denotes the predicted label for video $q$. Notably, the original Protonet calculates the Euclidean distances, we find that cosine
	distances performs better under our setting.

	\section{Experiments}
	\subsection{Datasets}
	To verify the effectiveness of our method, we conduct experiments on three widely used detests for video action recognition.
	
	\noindent\textbf{Kinetics} 
	The original Kinetics dataset \cite{kay2017kinetics} contains 400 action classes and 306,245 videos. We follow the splits constructed by CMN \cite{zhu2018compound}, which contains 100 action classes, each with 100 videos. 
	64, 12 and 24 non-overlapping classes are used as training set, validation set, and testing set, respectively.
	
	\noindent\textbf{UCF101} 
	UCF101 dataset \cite{soomro2012ucf101} has 101 action categories and 13,320 video clips. 
	We follow the splits proposed in \cite{zhang2020few}, in which 70, 10, 21 disjoint action categories are sampled for training, validation and testing.
	
	\noindent\textbf{HMDB51} 
	HMDB51 dataset \cite{kuehne2011hmdb} has 51 action classes and 6,849 videos in total.  
	We follow the same split as \cite{zhang2020few}. 
	31, 10, 10 action classes are selected for training, validation and testing, respectively.

	\subsection{Implementation details}
	\noindent\textbf{Pre-processing.} 
	As described in section \ref{section:shift}, for each video, we sample RGB and depth clips of length $num_{seg} * num_{f}$ which are used to extract RGB feature and depth feature. In our experiments, both $num_{seg}$ and $num_{f}$ are set to 4. During the meta-train phase, the $num_{aug}$ is set to 2, while during the meta-test phase we only sample the strictly matched $\left \langle RGB\ clip,\ depth\ clip \right \rangle$ pair from the middle of all the segments. As for the processing of images, we first resize them to $256 \times 256$, and then randomly crop a $224 \times 224$ region. In the testing phase, we adopt a center crop to obtain the $224 \times 224$ region.
	
	\begin{table*}[!h]
		\caption{Classification accuracy (\%) of 5-way 1-shot, 2-shot, 3-shot, 4-shot, and 5-shot video recognition on the Kinetics dataset. Comparing against RGB Basenet, RGB Basenet++ utilizes the basic sampling strategy (section~\ref{section:shift}). Comparing against RGB + Depth + DGAdaIN, our full model AMeFu-Net utilizes the proposed temporal asynchronization augmentation.
			\label{tab:kinetics}}
		\begin{tabular}{c|c|c|c|c|c|c}
			\hline 
			& Model  & 1-shot  & 2-shot  & 3-shot  & 4-shot  & 5-shot \tabularnewline
			\hline 
			\hline 
			Baselines & RGB Basenet & 65.1 & 74.9 & 78.9 & 81.2 & 81.9 \tabularnewline
			\cline{2-7} 
			& RGB Basenet++ &  66.5  & 77.2 & 80.7 & 82.9 & 83.7  \tabularnewline
			\cline{2-7} 
			\hline 
			\hline
			Competitors & CMN \cite{zhu2018compound} & 60.5 & 70.0 & 75.6 & 77.3 & 78.9 \tabularnewline
			\cline{2-7} 
			& ARN \cite{zhang2020few} & 63.7 & - & - & - & 82.4 \tabularnewline
			\cline{2-7} 
			& TRAN \cite{bishay2019tarn}     & 66.6 & 74.6 & 77.3 &  78.9  & 80.7 \tabularnewline
			\cline{2-7} 
			& CFA \cite{hu2019weakly}     & 69.9 & -  & 80.5 &  -   & 83.1 \tabularnewline
			\cline{2-7} 
			& Embodied Learning \cite{fu2019embodied}     & 67.8 & 77.8 & 81.1 &  82.6  & 85.0 \tabularnewline
			\cline{2-7} 
			&  TAM \cite{cao2019few}     & 73.0 & -  & - &  -  & 85.8 \tabularnewline
			\hline 
			\hline
			Ours & RGB + Depth + DGAdaIN &  73.6  & 80.7  & 84.0 & 85.4 &  86.6  \tabularnewline
			\cline{2-7} 
			& AMeFu-Net &  \textbf{74.1 }  & \textbf{ 81.1 } & \textbf{84.3} & \textbf{85.6} & \textbf{86.8} \tabularnewline
			\hline 
		\end{tabular}
		\par
	\end{table*}

	\noindent\textbf{Training details.} 
	Due to the large difference between RGB modality and depth modality, we employ a two-stage training procedure. We first train the RGB sub-model and the depth sub-model, the feature extractors described in section~\ref{backbone}, with the classical training method under the supervised learning setting separately. Concretely, we replace the last fully connected layer in the original ResNet-50 with our own fully connected layer as a classifier to construct the RGB and depth sub-models. For the RGB sub-model, we finetune it on the source data for 6 epochs, the learning rate of the Resnet-50 $lr_{1}$ is set to 0.0001, and the learning rate of our fully connected layer for classification $lr_{2}$ is set to 0.001. For the depth sub-model, since the feature of the depth frames is hard to extract, we finetune it for 60 epochs. The learning rates $lr_{1}$ and $lr_{2}$ are both set to 0.0001, and we reduce it by 0.1 after 30 epochs. These pre-trained sub-models are further used as the feature extractors which extract the 2048-dimension features for RGB and depth, respectively. 
	
	We then use the depth guided adaptive fusion module (DGAdaIN) to fuse the features from these two streams. The DGAdaIN is finetuned in the meta-learning way for another 6 epochs, and the episodes $episode_{train}$ is set to 2000 every epoch. What is worth mentioning is that we train the DGAdaIN module under the 5-way-1-shot setting. The learning rate of DGAdaIN module $lr_{3}$ is set to 0.00002 and the parameters of two sub-models are not updated. 
	
	In addition, for all the experiments, we set the batch size to 6.
	Stochastic Gradient Descent (SGD) with momentum=0.9 is selected as our optimizer. 
	For UCF101 and HMDB51, we train the RGB sub-model and depth sub-model under the same setting with Kinetics. However,  considering the scale of the datasets is relatively small, we set $lr_{3}$ to 0.00001, $episode_{train}$ to 1000, and reduce the $lr_{3}$ by 0.1 after 3 epochs.

	\noindent\textbf{Evaluation.} During the meta-test phase, we randomly sample 10,000 episodes and the average accuracy is reported. Notably, We implement $L_2$ normalization on the fused features before they are fed into our few-shot classifier.

	\subsection{Results on Kinetics}
	\noindent\textbf{Baselines and competitors.} 
	We compare our method with several baselines to show the effectiveness of our model. We conduct the experiments under $5-way-K-shot$ setting, and we report 1-shot, 2-shot, 3-shot, 4-shot and 5-shot results in Tab.~\ref{tab:kinetics}. (1) For the first baseline "RGB Basenet", we follow the setting of "Basenet + test" reported in Embodied Learning \cite{fu2019embodied}, which adopts an ImageNet pre-trained ResNet-50 as RGB feature extractor directly and uses the Protonet as the few-shot classifier.  (2) For the second baseline "RGB Basenet++", we also directly use the Imagenet pre-trained ResNet-50 as our RGB feature extractor. The differences between these two baselines lie in two aspects: a) "RGB  Basenet" samples 16 consecutive frames, while "RGB Basenet++" samples $4*4$ frames as stated above. b) "RGB Basenet" uses the original Protonet, while we change Euclidean distance to cosine similarity. Besides, these baselines are all conducted without the temporal asynchronization augmentation mechanism.
	
	We also compare our method with several state-of-the-art works. (1) CMN \cite{zhu2018compound} mainly proposes a memory network structure. (2) TRAN \cite{bishay2019tarn} learns to compare the representation of variable temporal sequence, and calculates the relation scores between the query video and support videos. (3) CFA \cite{hu2019weakly} mainly emphasizes the importance of compositionality in visual concept representation and proposes the compositional feature aggregation (CFA) module. (4) Embodied Learning \cite{fu2019embodied} learns from virtual videos generated by embodied agents and further proposes a video segment augmentation method. (5) TAM \cite{cao2019few} proposes a temporal alignment algorithm to measure the similarities between query video and support videos. All the competitors conduct experiments with the same splits of Kinetics.
	
	For ours, the results of "RGB + depth + DGAdaIN" and "AMeFu-Net" are reported. The only difference between these two models is whether the temporal asynchronization mechanism is applied. More specifically, for "RGB + depth + DGAdaIN", we adopt the DGAdaIN module to fuse the RGB feature and depth feature, however, we just use the matched RGB and depth pair to train our model. While "AMeFu-Net" means the full model.
	
	\noindent\textbf{Results.} 
	From the results shown in Tab.~\ref{tab:kinetics}, we make the following observations:
	(1) Our method outperforms all the baselines and competitors, establishing new state-of-the-art with 74.1\%, 84.3\% and 86.8\% on 1-shot, 3-shot and 5-shot settings, respectively. (2) We notice that the strong baseline "RGB Basenet++" is superior to "RGB Basenet", which is mainly contributed by our basic sampling strategy (section~\ref{section:shift}). 
	It quite makes sense, since the clips of length $num_{seg} * num_{f}$ sampled by our method capture more temporal information, especially in the case of long-term videos. 
	We also observe that replacing the Euclidean to cosine distance in the Protonet classifier contributes a little to the performance.
	(3) The effect of our temporal asynchronization augmentation mechanism is shown by comparing the result of "RGB + depth + DGAdaIN" with "AMeFu-Net".
	 The performance is improved consistently for all shots which verifies that our model is more robust with the augmented video feature representation.
	(4) Generally, the performance of all the models is getting better as the number of labeled samples increases. However, the improvement brought by our method is getting smaller. This phenomenon shows that our method is most effective when data are extremely scarce.

\begin{figure*}[h]
		\centering
		{\includegraphics[width=0.9\linewidth]{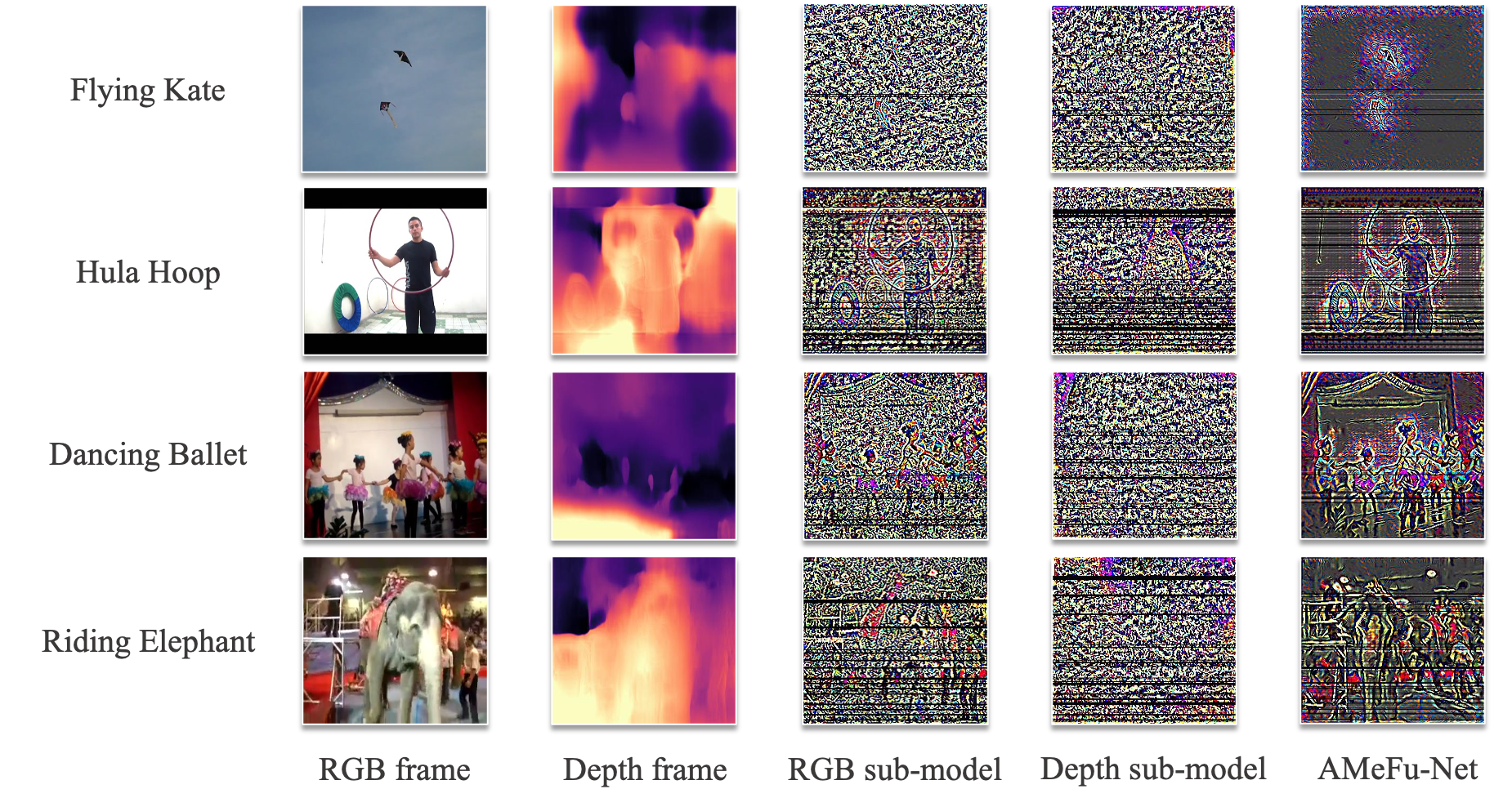}}
		\vspace{-0.15in}
		\caption{Class Activation Map (CAM) of our models. We visualize 4 examples sampled from Kinetics datasets. The first and second columns show the original RGB frame and the corresponding depth frame. The CAM of RGB sub-model, depth sub-model, and AMeFu-Net (ours) are displayed in the third, fourth and fifth columns, respectively. Best viewed in color. \label{visulization} 
		}
	\end{figure*}

	\noindent\textbf{Visualization.} We visualize the Class Activation Map (CAM) \cite{zhou2016learning} of some examples sampled from the validation and testing sets of Kinetics dataset using Grad-CAM \cite{selvaraju2017grad}. As shown in Fig.~\ref{visulization}, the third and fourth columns show the class activation maps computed from the RGB frames (first column) and depth frames (second column) individually. The fifth column shows the CAM of our AMeFu-Net, in which both the RGB clip and depth clip are fed into our model. In particular, the gradients of the action with the highest probability are back-propagated to highlight its map of interest. 
	
	We see that the RGB sub-model outperforms the depth sub-model, which is consistent with the fact that the RGB modality contains most of the visual information. The most important regions are recognized by the RGB sub-model, such as the person and the hula hoop in the example of "Hula Hoop", and the "kids" in the example of "Dancing Ballet". Although the CAM of depth sub-model does not seem as good as that of the RGB sub-model, the performance of RGB sub-model is improved by introducing the depth modality, which is demonstrated in the fifth column. Comparing the result of the “RGB sub-model" with "AMeFu-Net", we note that AMeFu-Net not only recognizes the important regions correctly but also eliminates the noises brought by irrelevant background to a large extent.

	\subsection{Results on UCF101 and HMDB51}
	
	\noindent \textbf{Baselines and Competitors.} 
	We also conduct experiments on UCF101 and HMDB51.
	We report the 1-shot, 3-shot, 5-shot results on both datasets.
	As for baselines, the "RGB Basenet", "RGB Basenet++" are reported.  
	The baselines and their settings are the same as those on Kinetics.

	\begin{table*} 
		\centering{}{\caption{Classification accuracy (\%) of 5-way 1-shot, 3-shot and 5-shot video recognition on UCF101 and HMDB51. 
				Comparing against RGB Basenet, RGB Basenet++ utilizes our basic sampling strategy (section~\ref{section:shift}).
				\label{tab:ucf-and-hmdb}}
		}%
		\begin{tabular}{c|c|c|c|c|c|c|c}
			\hline 
			& \multicolumn{1}{c|}{} & \multicolumn{3}{c|}{UCF101} & \multicolumn{3}{c}{HMDB51}\tabularnewline
			\hline 
			& Model & 1-shot & 3-shot & 5-shot & 1-shot & 3-shot & 5-shot\tabularnewline
			\hline 
			{Baselines} & RGB Basenet & 76.4 & 88.7 & 92.1 & 48.8 & 62.4 & 67.8\tabularnewline
			\cline{2-8} \cline{3-8} \cline{4-8} \cline{5-8} \cline{6-8} \cline{7-8} \cline{8-8} 
			& RGB Basenet++ & 78.5 & 90.0 & 92.9 & 49.3 & 63.0 & 68.2 \tabularnewline
			\cline{2-8} \cline{3-8} \cline{4-8} \cline{5-8} \cline{6-8} \cline{7-8} \cline{8-8} 
			
			\hline 
			{Competitors} & ARN~\cite{zhang2020few} & 66.3 & - & 83.1 & 45.5 & - & 60.6\tabularnewline
			\hline 
			\hline 
			Ours & AmeFu-Net & \textbf{85.1} & \textbf{93.1} & \textbf{95.5} & \textbf{60.2} & \textbf{71.5} & \textbf{75.5}\tabularnewline
			\hline 
		\end{tabular}
	\end{table*}

	\noindent \textbf{Results.} 
	Our results are shown in Tab.~\ref{tab:ucf-and-hmdb}. 
	Our method achieves state-of-the-art performance on all settings. 
	For UCF101, we achieve 85.1\% on 1-shot and 95.5\% on 5-shot, outperforming the ARN \cite{zhang2020few} by 18.8\% and 12.4\%, respectively. 
	For HMDB51, our method achieves 60.2\% on 1-shot, 71.5\% on 3-shot and 75.5\% on 5-shot, outperforming the ARN \cite{zhang2020few} by 14.7\% on 1-shot setting. 
	We also want to highlight some other results. 
	First, we notice the performance improvement by comparing "RGB Basenet++" against "RGB Basenet". 
	It keeps consistent with the results on the Kinetics dataset. 
	Second, the results on UCF101 are much better than those on HMDB51. 
	Take the 1-shot as an example, our model achieves 85.1\% on UCF101 while HMDB51 only reaches 60.2\%.  
	Notably, the action classes of UCF101 are much easier to recognize, such as "Apply Eye Makeup", "Baby Crawling" and "Sky Diving". These actions are not difficult to be distinguished. In some extreme cases, we can recognize the actions even with only one keyframe. In comparison, HMDB51 is much more challenging, considering some facial actions are contained, such as "smile", "laugh", "chew" and "talk". Such actions increase the difficulty of video recognition a lot.

	\subsection{Ablation study}
	To verify the effectiveness of components contained in our model, we implement experiments to evaluate them on the Kinetics dataset. Generally, we conduct experiments on the 5-way setting and report the result of 1-shot to 5-shot. The results are reported in Tab.~\ref{tab:abalation}.

	\begin{table}\small
		\begin{centering}
			\caption{Ablation studies on fusion strategies, fusion modules, and temporal shifting times. 
				Classification accuracy (\%)  of 5-way few-shot on Kinetics are reported.
				\label{tab:abalation}}
			\begin{tabular}{c|c|c|c|c|c}
				\hline
				Setting & 1-shot  & 2-shot  & 3-shot  & 4-shot  & 5-shot \tabularnewline
				\hline 
				\hline 
				&  \multicolumn{5}{c}{fusion strategies}  \tabularnewline
				\hline
				RGB +  Depth + Concat &  67.5  &  76.9 & 81.1 &  82.0 &   83.3 \tabularnewline	
				\hline
				\cline{2-6} 
				RGB + Depth + DGAdaIN &  \textbf{73.6}  & \textbf{80.7}  & \textbf{84.0} & \textbf{85.4} &  \textbf{86.6}  \tabularnewline
				\hline 
				\hline
				& \multicolumn{5}{c}{fusion modules}  \tabularnewline
				\hline 
				RGB guide Depth & 50.6  & 58.7  & 64.3  & 64.1   & 66.1 \tabularnewline
				\cline{2-6}
				\hline
				DGAdaIN & \textbf{73.6}  & \textbf{80.7}  & \textbf{84.0}  & \textbf{85.4}   & \textbf{86.6} \tabularnewline
				\cline{2-6}
				\hline
				two-way guidance  & 66.5  & 74.8  & 78.3  & 79.4  & 82.8 \tabularnewline
				\cline{2-6}
				\hline
				\hline
				& \multicolumn{5}{c}{temporal shifting times}  \tabularnewline
				\hline
				$num_{aug} =1 $ & 73.0  & 80.6  & \textbf{84.5} & 85.5  & 86.0 \tabularnewline
				\cline{2-6}
				\hline
				$num_{aug} =2$ & \textbf{74.1}  & \textbf{81.1}  & 84.3  & \textbf{85.6}   & \textbf{86.8} \tabularnewline
				\cline{2-6}
				\hline
				$num_{aug} =3 $ & 72.1  & 80.5 & 84.4  & 85.4  & 86.0 \tabularnewline
				\hline
				\hline
			\end{tabular}
			\par\end{centering}
	\end{table}

	\noindent\textbf{How to fuse RGB stream and depth stream?} The most natural way to fuse multi-modality features is to concatenate different features directly. 
	We implement this naive method which termed as "RGB + Depth + Concat". 
	Specifically, we first concatenate the RGB feature extracted by the "RGB sub-model" and the depth feature extracted by the "depth sub-model", and then the concatenated feature is fed into a fully connected layer. 
	The training procedure is the same with ours "RGB + Depth + DGAdaIN" which is not equipped with the temporal asynchronization augmentation mechanism. 
	The results in Tab.~\ref{tab:abalation} show that our method achieves superior performance.

	\noindent\textbf{How to design the adaptive fusion module?} 
	DGAdaIN deforms the RGB feature towards depth feature.
	We also explore other strategies.
	For "RGB guide Depth", we learn the affine parameters from the RGB feature and use them to deform depth feature. 
	For "two-way guidance", we average the output feature of "RGB guide depth" and "depth guide RGB" and then feed it to the classifier.
	We fine-tune the different modules with the same setting.
	Results in Tab.~\ref{tab:abalation} show that our DGAdaIN achieves the best performance, outperforming the worst "RGB guide depth" by a large margin. 
	As analyzed above, the most important visual context such as main objects can still be extracted from the RGB frames, so we have to keep the "context" unchanged while use the depth as the "style" to guide the original RGB features.  
	
	\noindent\textbf{How many times should we shift depth?} 
	We explore three different settings of the $num_{aug}$. 
	Results in Tab.~\ref{tab:abalation} show that as the $num_{aug}$ increases, the performance of our model will finally converge. In most cases, our model achieves the best performance when the $num_{aug}$ is set to 2. 
	That is because when the $num_{aug}$ is set to 1, which means we only shift once, the performance of augmentation is limited. 
	When we set the $num_{aug}$ $\textgreater$ 2, training on $num_{aug}$ un-matched RGB and depth pairs may bring too much misleading information. In such cases, the depth clips may cause undesired noises. We set $num_{aug}$ to 2 in our final model.

	\section{Conclusion} 
	In this paper, we have explored and exploited the depth information to alleviate the extremely data-scarce problem in few-shot video action recognition. Technically, we have proposed a novel temporal asynchronization augmentation strategy to augment source video representation. In addition, we have proposed a depth guided adaptive instance normalization module (DGAdaIN) to fuse the features from two different streams adaptively. Extensive experiments show that our proposed method established new state-of-the-art on three widely used datasets including Kinetics, UCF101, and HMDB51.

	\bibliographystyle{ACM-Reference-Format} 
	\bibliography{references}
\end{document}